\documentclass[submission,copyright,creativecommons]{eptcs}
\providecommand{\event}{SOS 2007} 
\usepackage{breakurl}             
\usepackage[utf8]{inputenc}

\usepackage{color}
\usepackage{amsmath}
\usepackage{amsthm}
\usepackage{amssymb}
\usepackage{graphicx}

\usepackage{url}

\newtheorem{theorem}{Theorem}[section]
\newtheorem{prop}[theorem]{Proposition}

\title{Enclosing the Sliding Surfaces of a Controlled Swing}
\author{Luc Jaulin
\institute{ENSTA-Bretagne\\ Brest, France}
\institute{Robex, Lab-STICC\\
}
\email{lucjaulin@gmail.com}
\and
Benoît Desrochers
\institute{DGA-TN\\
Brest, France}
\email{\quad benoit.desrochers@ensta-bretagne.org}
}
\def\titlerunning{Enclosing the Sliding Surfaces of a Controlled Swing}
\def\authorrunning{J. Jaulin \& B. Desrochers}
\begin{document}
\maketitle

\begin{abstract}
When implementing a non-continuous controller for
a cyber-physical system, it may happen that the evolution of the closed-loop
system is not anymore piecewise differentiable along the trajectory,
mainly due to conditional statements inside the controller. This may
lead to some unwanted chattering effects than may damage the system.
This behavior is difficult to observe even in simulation. In this
paper, we propose an interval approach to characterize the \emph{sliding
surface} which corresponds to the set of all states such that the
state trajectory may jump indefinitely between two distinct behaviors.
We show that the recent notion of \emph{thick sets} will allows us
to compute efficiently an outer approximation of the sliding surface
of a given class of hybrid system taking into account all set-membership
uncertainties. An application to the verification of the controller
of a child swing is considered to illustrate the principle of the
approach.
\end{abstract}

\section{Introduction}

The verification of the properties of cyber-physical systems \cite{taha:13:zeno,taha:15:acumen}
is a fundamental problem for which set membership techniques have
provided original and efficient results \cite{rohouAut18} \cite{Rauh2009IntervalAT}.

Different types of such approaches have been studied for the verification.
Some require the integration of nonlinear differential equations \cite{alexandre:chap:16}\cite{wilczak2011}\cite{Mitchell07}.
Others are based on positive invariance approaches 
\cite{Asarin07}
\cite{lemezo:tac:18}.
For the numerical resolution some methods build a grid of the state
space \cite{SaintPierre02}\cite{Delanoue:attraction:06} which makes
them computationally expensive. Lyapunov-based methods \cite{Ratschan:She:10},
level-set methods \cite{mitchell:validating:2001}, or barrier functions
\cite{Bouissou14} are attractive, since they do not perform any integration
through time. Now, these methods generally require a parametric expression
for candidate Lyapunov-like functions which is not always realistic.

This paper considers the verification of controlled cyber-physical
systems \cite{Ramdani:Nedialkov11} which include both a physical
system and a control algorithm. The verification requires approaches
coming from invariance approaches \cite{Blanchini08}, static analysis
\cite{Goubault06staticanalysis} and abstract interpretation \cite{Cousot97}.

To detect the discontinuities, we propose in this
paper to characterize the set of states around which undesirable switching
phenomena could occur. The corresponding zone is called a {\emph{sliding
surface} which may become thick in case of uncertainties.
In practice, the system can be trapped inside the sliding surface
without any possibility to escape.

In \cite{jaulin:sliding:20}\cite{Romig19}, it
has been shown that sliding surfaces can be characterized rigorously
using interval techniques \cite{Moore79}\cite{Kreinovich:97} for
hybrid systems without any uncertainties. In this paper, we extend
this approach to uncertain hybrid systems.

The paper is organized as follows. Section \ref{sec:formalism} provides
the formalism and defines sliding surfaces. Section \ref{sec:Thick-sets}
introduces \emph{thick sets} that will allow us to extend the concept
of sliding surfaces to the case of uncertainty. Section \ref{sec:application}
shows how our approach can be used to validate the controller of a
child swing. Section \ref{sec:Conclusion} concludes the paper.

\section{Formalism\label{sec:formalism}}

\subsection{Hybrid system}

In this paper, we consider a specific class of hybrid dynamical systems
of the form 
\begin{equation}
\mathcal{S}\left(\mathbb{A}\right):\begin{array}{c}
\left\{ \begin{array}{ccc}
\dot{\mathbf{x}}=\mathbf{f}_{a}\left(\mathbf{x}\right) &  & \text{if \ensuremath{\mathbf{x}\in\mathbb{A}}}\\
\dot{\mathbf{x}}=\mathbf{f}_{b}\left(\mathbf{x}\right) &  & \text{if \ensuremath{\mathbf{x}\in\mathbb{B}=\overline{\mathbb{A}}}}
\end{array}\right.\end{array}\label{eq:formal}
\end{equation}
where
\begin{itemize}
\item $\mathbf{f}_{a},\mathbf{f}_{b}$~:$\mathbb{R}^{n}\rightarrow\mathbb{R}^{n}$
are continuous and differentiable,
\item $\mathbb{A}$ is a closed subset of $\mathbb{R}^{n}$ that can be
defined by inequalities linked by Boolean operators.
\end{itemize}
This definition is illustrated by the automaton of Figure \ref{fig:automaton},
taking the conventions used for hybrid systems \cite{Asarin02,Frehse:08}.
The red arrows show transitions which may generate the sliding phenomena
that are studied in this paper.

\begin{figure}
\centering\includegraphics[width=0.4\textwidth]{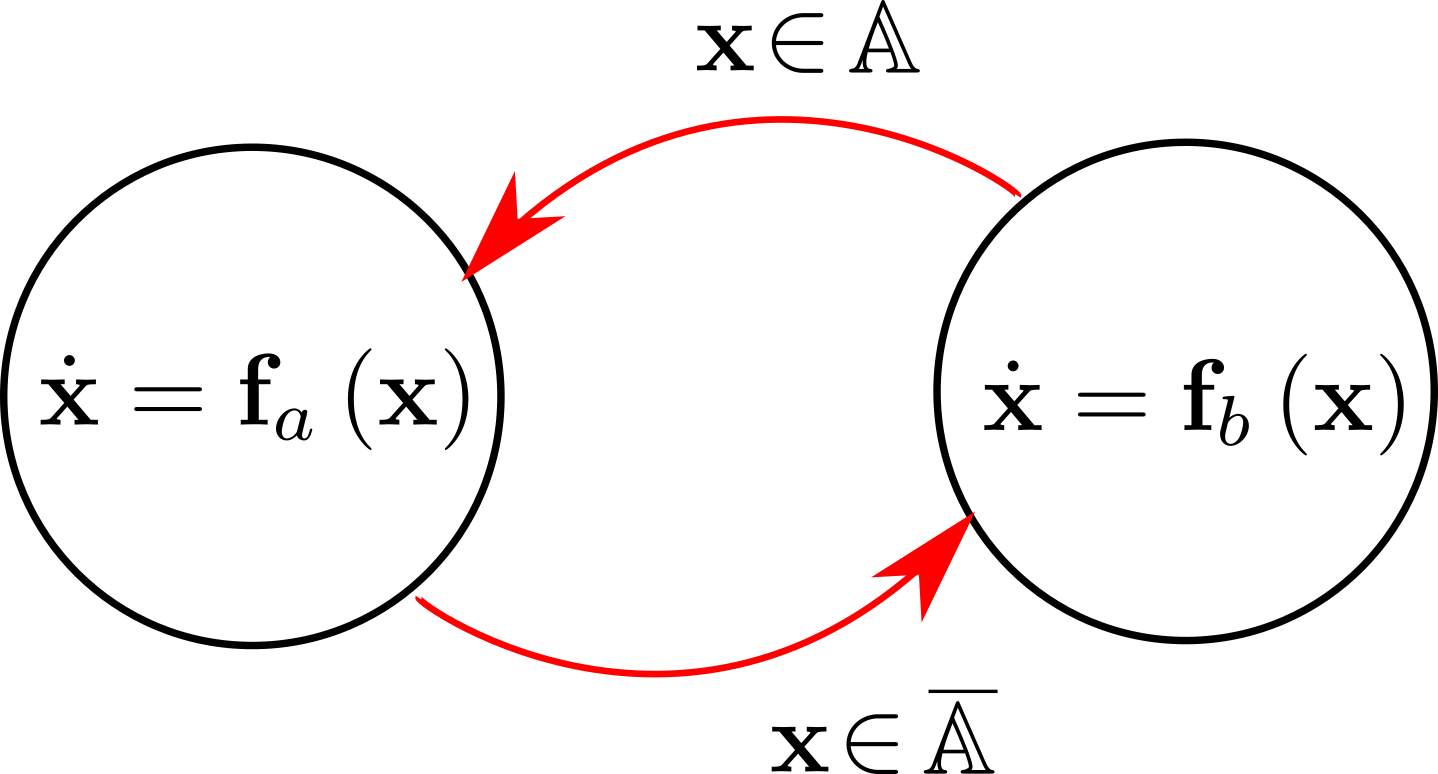}
\caption{Automaton representing our Cyber Physical System}
\label{fig:automaton}
\end{figure}

\subsection{Algebra of closed sets \label{subsec:Algebra-of-closed}}

Given a set $\mathbb{X}\subset\mathbb{R}^{n}$, we denote by $\text{cl}(\mathbb{X})$
the smallest closed subset of $\mathbb{R}^{n}$ which contains $\mathbb{X}$.
The intersection of closed sets and the finite union of closed sets
if still a closed set. We define the \emph{closed complementary} $\overline{\mathbb{A}}$
of the closed set $\mathbb{A}$ as 
\[
\overline{\mathbb{A}}=\text{cl}\{\mathbf{x}|\mathbf{x}\notin\mathbb{A}\}.
\]
The \emph{boundary} of the closed set $\mathbb{A}$ is denoted by
$\partial\mathbb{A}$. The closed set $\mathbb{A}$ is \emph{topologically
stable} if $\partial\mathbb{A}=\partial\mathbb{\overline{A}}.$

For instance, the disk of $\mathbb{D}=\{\mathbf{x}\in\mathbb{R}^{2}|\|\mathbf{x}\|\leq1\}$
is topologically stable with $\overline{\mathbb{D}}=\{\mathbf{x}\in\mathbb{R}^{2}|\|\mathbf{x}\|\geq1\}$
and $\partial\mathbb{D}=\partial\mathbb{\overline{D}}=\{\mathbf{x}\in\mathbb{R}^{2}|\|\mathbf{x}\|=1\}$.
But the circle $\mathbb{C}=\{\mathbf{x}\in\mathbb{R}^{2}|\|\mathbf{x}\|=1\}$
is not topologically stable. Indeed $\overline{\mathbb{C}}=\text{cl}\{\mathbf{x}\in\mathbb{R}^{2}|\|\mathbf{x}\|\neq1\}=\mathbb{R}^{2}$
and $\partial\mathbb{C}=\{\mathbf{x}\in\mathbb{R}^{2}|\|\mathbf{x}\|=1\}\neq\partial\mathbb{\overline{C}}=\emptyset$. 

In this paper, we will assume that 
\begin{enumerate}
\item closed sets $\mathbb{A}$ involved in the formulation (\ref{eq:formal})
are topologically stable, \emph{i.e.}, they have the same boundary
as their interior. 
\item the closed sets can be defined as a finite composition (with unions,
intersections) of sets of the form 
\[
\mathbb{A}=\left\{ \mathbf{x}\in\mathbb{R}^{n}\,|\,c(\mathbf{x})\leq0\right\} 
\]
 where $c$ is a smooth function.
\end{enumerate}

\subsection{Lie derivative}

We recall the notion of Lie derivative that will be used to define
the sliding surfaces. Consider a function $c:\mathbb{R}^{n}\rightarrow\mathbb{R}$.
The \emph{Lie derivative} of $c$ with respect to the field $\mathbf{f}:\mathbb{R}^{n}\rightarrow\mathbb{R}^{n}$
as 
\begin{equation}
\mathcal{L}_{\mathbf{f}}^{c}\left(\mathbf{x}\right)=\frac{dc}{d\mathbf{x}}\left(\mathbf{x}\right)\cdot\mathbf{f}\left(\mathbf{x}\right).
\end{equation}
We also define the \emph{Lie set} as 
\begin{equation}
\mathbb{L}_{\mathbf{f}}^{c}=\left\{ \mathbf{x}|\mathcal{L}_{\mathbf{f}}^{c}\left(\mathbf{x}\right)\leq0\right\} .
\end{equation}
In our context, the field depends on $i\in\{a,b\}$ (see \ref{eq:formal}).
We will write 
\begin{equation}
\begin{array}{ccc}
\mathcal{L}_{i}^{c}\left(\mathbf{x}\right) & = & \mathcal{L}_{\mathbf{f}_{i}}^{c}\left(\mathbf{x}\right)\\
\mathbb{L}_{i}^{c} & = & \mathbb{L}_{\mathbf{f}_{i}}^{c}
\end{array}
\end{equation}

\subsection{Sliding surface}

The \emph{sliding surface }$\mathbb{S}\left(\mathbb{A}\right)$ \cite{Drakunov92}
for $\mathcal{S}\left(\mathbb{A}\right)$ (see Equation \ref{eq:formal})
is defined as the largest subset (with respect to the inclusion $\subset$)
of the boundary $\partial\mathbb{A}$ of $\mathbb{A}$ such that the
system can stay inside for a non-degenerate interval of time.

If $\mathbb{A}$ is defined by the inequality $c\left(\mathbf{x}\right)\leq0$,
then $\mathbb{B}=\overline{\mathbb{A}}$ is defined by $c\left(\mathbf{x}\right)\geq0$
and the boundary by $c\left(\mathbf{x}\right)=0$ (see Subsection
\ref{subsec:Algebra-of-closed}). The sliding surface is 
\begin{equation}
\begin{array}{ccc}
\mathbb{S}\left(\mathbb{A}\right) & = & \partial\mathbb{A}\cap\left\{ \mathbf{x}\,|\,\mathcal{L}_{a}^{c}\left(\mathbf{x}\right)\geq0\,\wedge\mathcal{L}_{b}^{c}\left(\mathbf{x}\right)\leq0\right\} \\
 & = & \partial\mathbb{A}\cap\overline{\mathbb{L}_{a}^{c}}\,\cap\mathbb{L}_{b}^{c}.
\end{array}\label{eq:SA}
\end{equation}

Figure \ref{fig:thA1.png}, taken from \cite{jaulin:sliding:20},
illustrates the principle of this proposition in the case where $\mathbb{A}$
is described by one inequality $c\left(\mathbf{x}\right)\leq0$. The
boundary $\partial\mathbb{A}$ of $\mathbb{A}$ is composed of four
parts : 
\[
\begin{array}{ccl}
\partial\mathbb{A}\cap\overline{\mathbb{L}_{a}^{c}\left(q\right)}\,\cap\overline{\mathbb{L}_{b}^{c}\left(q\right)}\, & \rightarrow & \text{magenta}\\
\partial\mathbb{A}\cap\overline{\mathbb{L}_{a}^{c}\left(q\right)}\,\cap\mathbb{L}_{b}^{c}\left(q\right)\, & \rightarrow & \text{red}\\
\partial\mathbb{A}\cap\mathbb{L}_{a}^{c}\left(q\right)\,\cap\mathbb{L}_{b}^{c}\left(q\right) & \rightarrow & \text{yellow}\\
\partial\mathbb{A}\cap\mathbb{L}_{a}^{c}\left(q\right)\,\cap\overline{\mathbb{L}_{b}^{c}\left(q\right)} & \rightarrow & \text{black}
\end{array}
\]
One trajectory (dotted line) $\mathbf{x}(t)$ is also represented.
Before the yellow arc, $c\left(\mathbf{x}\right)$ is positive and
decreases. When it crosses the yellow arc, $c\left(\mathbf{x}\right)=0$
for some isolated time point $t_{1}$. Then $\mathbf{x}(t)$ remains
inside $\mathbb{A}$ until it reaches the red arc. It slides in the
red arc for some non-degenerate time interval. When $\mathbf{x}(t)$
reaches the magenta arc, it leaves $\mathbb{A}$.

\begin{figure}
\centering\includegraphics[width=0.4\paperwidth]{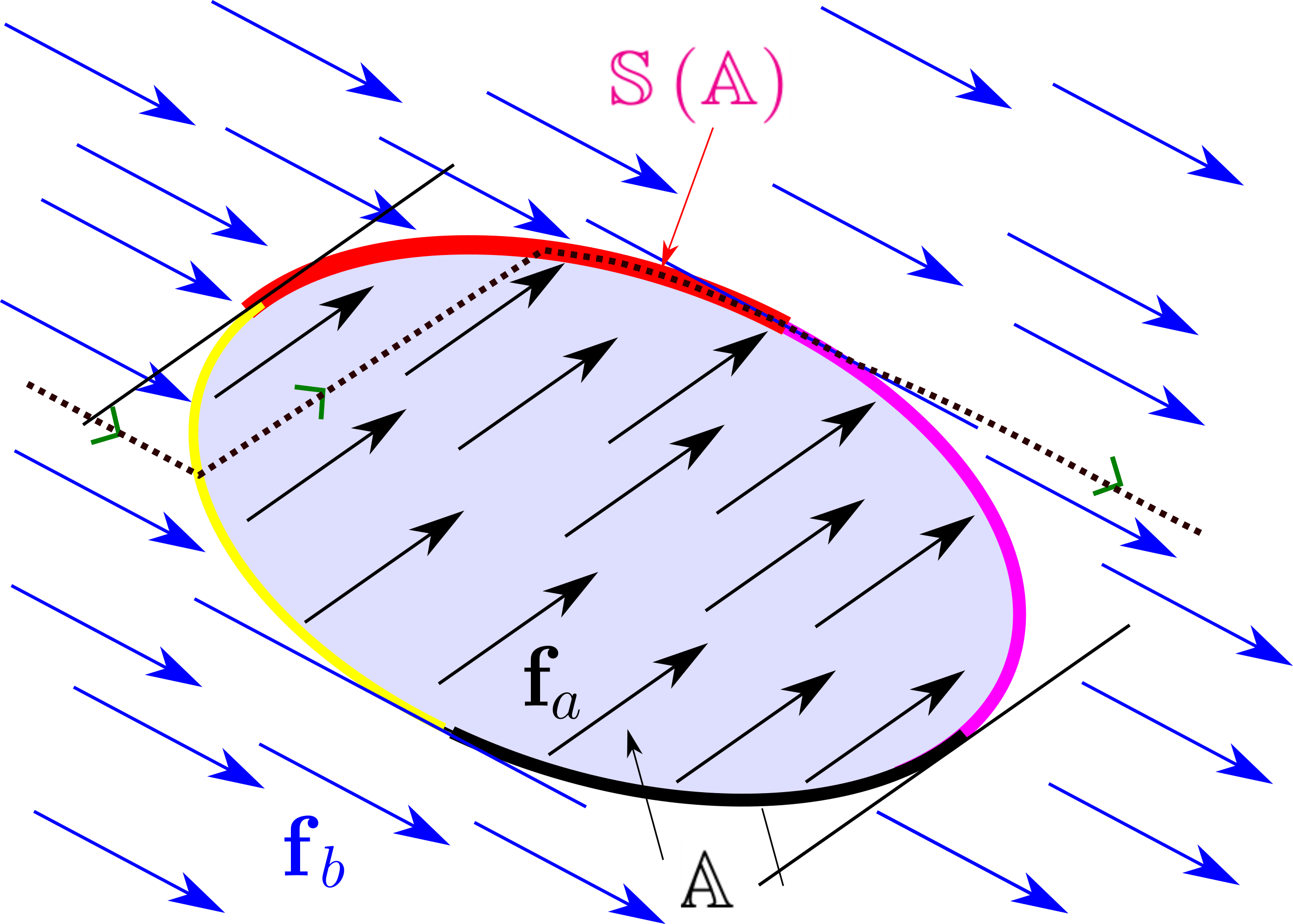}

\caption{Sliding surface $\mathbb{S}\left(\mathbb{A}\right)$ (red) for $\mathbb{A}=\left\{ \mathbf{x}|c\left(\mathbf{x}\right)\protect\leq0\right\} $}
\label{fig:thA1.png}
\end{figure}

We recall the following result that has been proved in \cite{jaulin:sliding:20}.

\begin{prop}
\label{prop:algebra}Consider two closed sets $\mathbb{A}_{1}$ and
$\mathbb{A}_{2}$. As illustrated by Figure \ref{fig:thA1A2}, we
have
\begin{equation}
\begin{array}{cccc}
(i) & \mathbb{S}\left(\mathbb{A}_{1}\cap\mathbb{A}_{2}\right) & = & \left(\mathbb{S}\left(\mathbb{A}_{1}\right)\cap\mathbb{A}_{2}\right)\cup\left(\mathbb{S}\left(\mathbb{A}_{2}\right)\cap\mathbb{A}_{1}\right)\\
(ii) & \mathbb{S}\left(\mathbb{A}_{1}\cup\mathbb{A}_{2}\right) & = & \left(\mathbb{S}\left(\mathbb{A}_{1}\right)\cap\overline{\mathbb{A}_{2}}\right)\cup\left(\mathbb{S}\left(\mathbb{A}_{2}\right)\cap\overline{\mathbb{A}_{1}}\right)
\end{array}
\end{equation}
\end{prop}

\begin{figure}
\centering\includegraphics[width=0.8\textwidth]{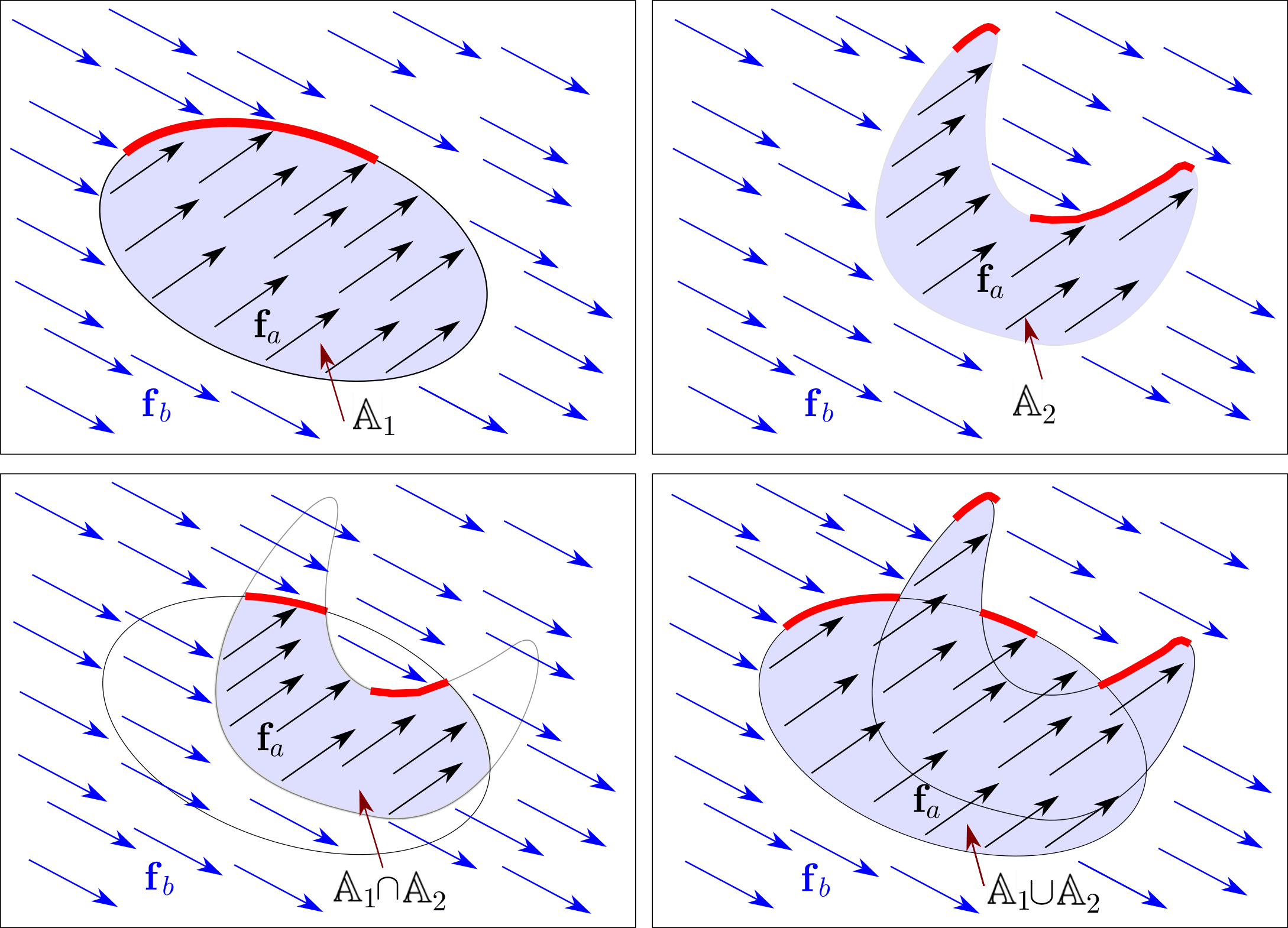}

\caption{Illustration of Proposition \ref{prop:algebra}, the sliding surfaces
are painted red}
\label{fig:thA1A2}
\end{figure}

Proposition \ref{prop:algebra} can be used to compute the sliding
surface of a set $\mathbb{A}$ as soon as $\mathbb{A}$ can be defined
by inequalities connected by Boolean operators such as \emph{and,
or, not.} The proposition is illustrated by Figure \ref{fig:thA1A2A3.png}
in the case where $\mathbb{A}=\mathbb{A}_{1}\cup\left(\mathbb{A}_{2}\cap\mathbb{A}_{3}\right)$
and $\mathbb{A}_{i}=\left\{ \mathbf{x}|c_{i}\left(\mathbf{x}\right)\leq0\right\} $.
The trajectory (green) slides twice, first on $\partial\mathbb{A}_{1}$,
then it slides on $\partial\mathbb{A}_{2}$. The sliding surfaces
are painted red.

\begin{figure}
\centering\includegraphics[width=0.6\textwidth]{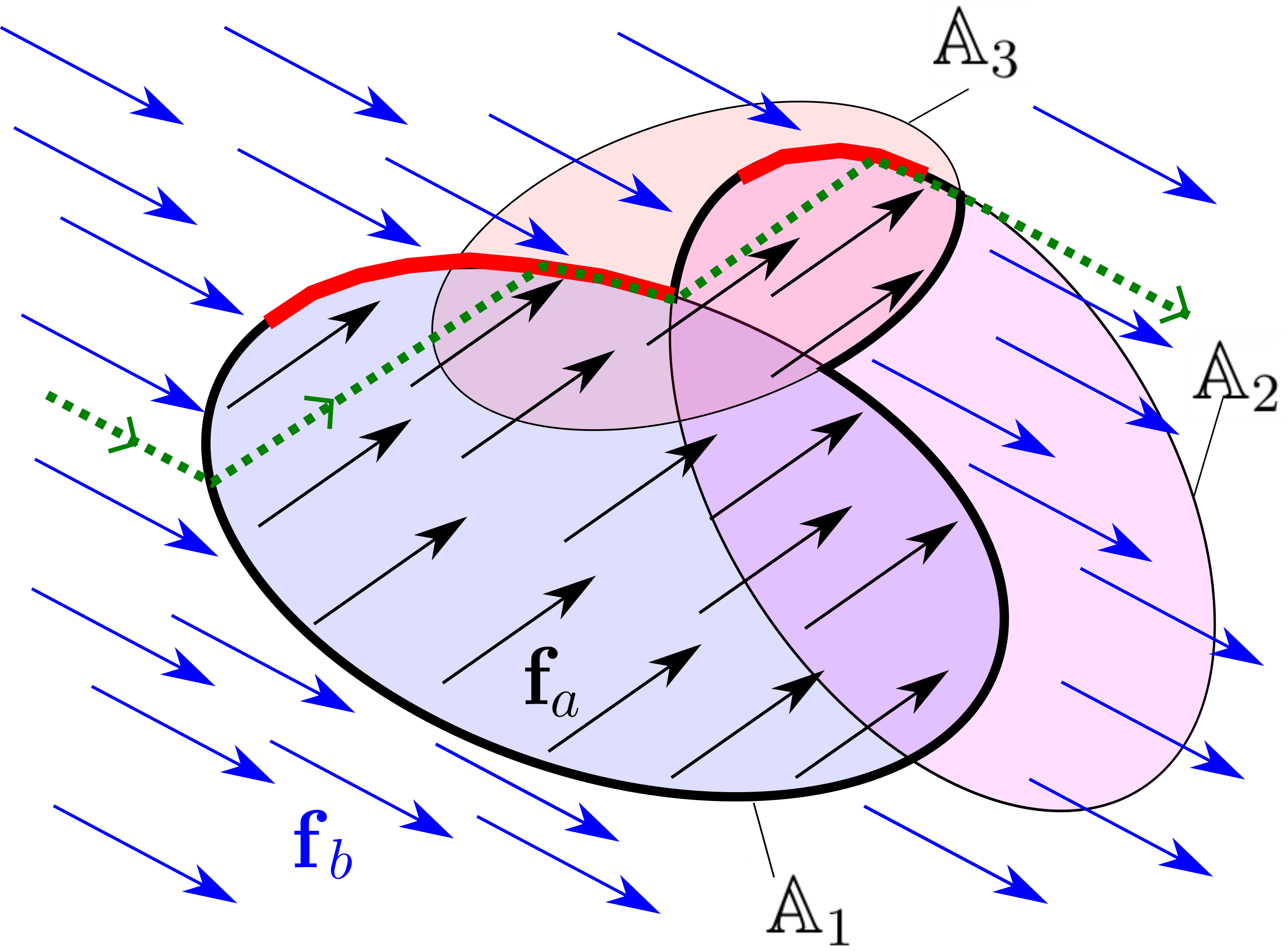}

\caption{Sliding surfaces for $\mathbb{A}=\mathbb{A}_{1}\cup\left(\mathbb{A}_{2}\cap\mathbb{A}_{3}\right)$}
\label{fig:thA1A2A3.png}
\end{figure}

\subsection{Sliding surface in case of uncertainties}

In case of uncertainties, the system defined in (\ref{eq:formal})
can be described as
\[
\begin{array}{c}
\mathcal{S}_{\mathbf{p}}:\begin{array}{c}
\left\{ \begin{array}{ccc}
\dot{\mathbf{x}}=\mathbf{f}_{a}\left(\mathbf{x}\right) &  & \text{if \ensuremath{\mathbf{x}\in\mathbb{A}}}\\
\dot{\mathbf{x}}=\mathbf{f}_{b}\left(\mathbf{x}\right) &  & \text{if \ensuremath{\mathbf{x}\in\overline{\mathbb{A}}}}
\end{array}\right.\end{array}\end{array}
\]
but now, $\mathbb{A},\mathbf{f}_{a},\mathbf{f}_{b}$ are uncertain.
For instance, they may depend on a parameter vector $\mathbf{p}\in[\mathbf{p}]$
which models the uncertainties. Recall that from Equation \ref{eq:SA},
the sliding surface satisfies 
\[
\begin{array}{ccl}
\mathbb{S} & = & \partial\mathbb{A}\cap\overline{\mathbb{L}_{a}^{c}}\,\cap\mathbb{L}_{b}^{c}\end{array}
\]
Assume that
\[
\left\{ \begin{array}{c}
\mathbb{A}^{\subset}\subset\mathbb{A}\subset\mathbb{A}^{\supset}\\
\mathbb{L}_{a}^{c\,\subset}\subset\mathbb{L}_{a}^{c}\subset\mathbb{L}_{a}^{c\,\supset}\\
\mathbb{L}_{b}^{c\,\subset}\subset\mathbb{L}_{b}^{c}\subset\mathbb{L}_{b}^{c\,\supset}
\end{array}\right.
\]
The sliding surface satisfies
\[
\underset{\mathbb{S}^{\subset}}{\underbrace{\partial\mathbb{A}^{\subset}\cap\overline{\mathbb{L}_{a}^{c\supset}}\,\cap\mathbb{L}_{b}^{c\subset}}}\subset\mathbb{S}\subset\underset{\mathbb{S}^{\supset}}{\underbrace{\partial\mathbb{A}^{\supset}\cap\overline{\mathbb{L}_{a}^{c\subset}}\,\cap\mathbb{L}_{b}^{c\supset}}}
\]
and we can write 
\[
\mathbb{S}\in[\negthinspace[\mathbb{S^{\subset}},\mathbb{S}^{\supset}]\negthinspace]=\partial[\negthinspace[\mathbb{A^{\subset}},\mathbb{A}^{\supset}]\negthinspace]\cap\overline{[\negthinspace[\mathbb{L}_{a}^{c\,\subset},\mathbb{L}_{a}^{c\,\supset}]\negthinspace]}\,\cap[\negthinspace[\mathbb{L}_{b}^{c\,\subset},\mathbb{L}_{b}^{c\,\supset}]\negthinspace]
\]
using the thick set formalism described in the following section.
In our application, is is clear that $\mathbb{S}^{\subset}=\emptyset$,
since the set to be enclosed is a surface that has no volume. The
set $\mathbb{S}^{\supset}$ is an upper approximation of this surface
as illustrated by Figure \ref{fig:slidingsetA}.

\begin{figure}
\centering\includegraphics[width=0.4\paperwidth]{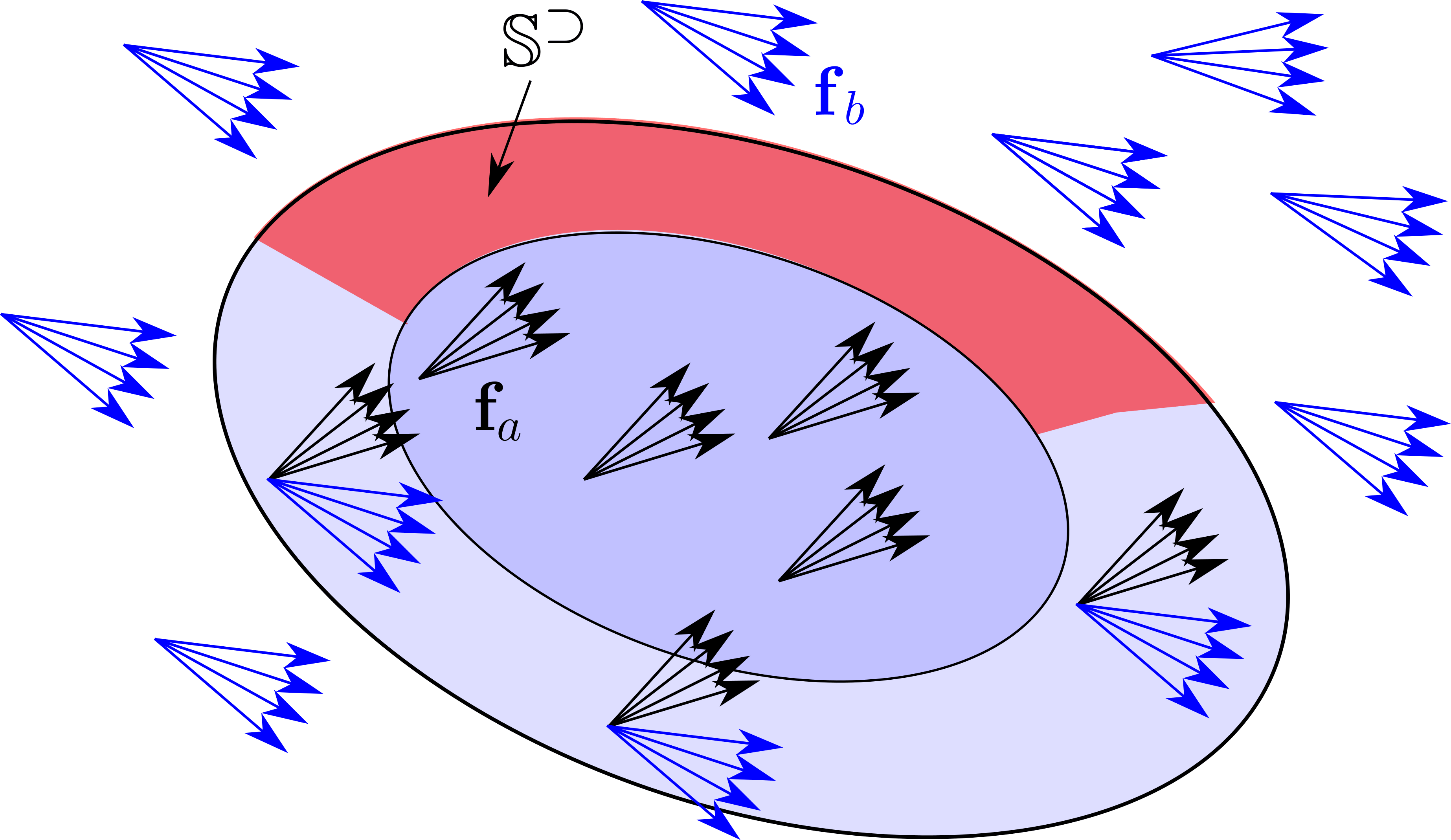}

\caption{The set $\mathbb{S}^{\supset}$ (red) is an upper approximation of
the sliding surface $\mathbb{S}_{\mathbf{p}}$ to be enclosed }
\label{fig:slidingsetA}
\end{figure}

\section{Thick sets\label{sec:Thick-sets}}

When dealing with uncertainties, the evolution equation involved in
Equation \ref{eq:formal} and the sets $\mathbb{A},\mathbb{B}$ become
uncertain. The sliding surface to be computed becomes thick and now
has an interior. To compute an inner and an outer approximation of
this approximation, we now introduce the recent concept of \emph{thick
set} with the associated algebra \cite{DesrochersThick2016}.

\subsection{Definition}

If an interval of $\mathbb{R}$ is an uncertain real number, a \emph{thick
set} is an uncertain subset of $\mathbb{R^{\textnormal{\ensuremath{n}}}}$.
More precisely, a thick set is an interval of the powerset of $\mathbb{R}^{n}$
equipped with the inclusion $\subset$ as an order relation.

A \emph{thin set} is a subset of $\mathbb{R^{\textnormal{\ensuremath{n}}}}$.
It is qualified as \emph{thin} because its boundary is thin. 

Denote by $(\mathcal{P}(\mathbb{R}^{n}),\subset)$, the powerset of
$\mathbb{R}^{n}$ equipped with the inclusion $\subset$ as an order
relation. A \emph{thick set} $[\negthinspace[\mathbb{X}]\negthinspace]$
of $\mathbb{R}^{n}$ is an interval of $(\mathcal{P}(\mathbb{R}^{n}),\subset)$.
If $[\negthinspace[\mathbb{X}]\negthinspace]$ is a thick set of $\mathbb{R}^{n}$
, there exist two subsets of $\mathbb{R}^{n}$ , called the \emph{subset
bound} and the \emph{superset bound} such that
\begin{equation}
[\negthinspace[\mathbb{X}]\negthinspace]=[\negthinspace[\mathbb{\mathbb{X}^{\subset}},\mathbb{X}^{\supset}]\negthinspace]=\{\mathbb{X}\in\mathcal{P}(\mathbb{R}^{n})|\mathbb{\mathbb{X}^{\subset}\subset\mathbb{X}\subset\mathbb{X}^{\supset}}\}.
\end{equation}

The subset $\mathbb{X}^{\supset}\backslash\mathbb{X}^{\subset}$ is
called the \emph{penumbra} and plays an important role in the characterization
of thick sets \cite{DesrochersTAC16}. Thick sets can be used to represent
uncertain sets (such as an uncertain map \cite{desrochers:iros2015})
or soft constraints \cite{Brefort14}. It can also can also be extended
to fuzzy sets as shown in \cite{dubois:jaulin:prade}\cite{redaTSF20}. 

\subsection{Algebra}

We now show how we can define operations for thick sets (as a union,
intersection, difference, etc.). The main motivation is to be able
to compute with thick sets.

Consider two thick sets $[\negthinspace[\mathbb{X}]\negthinspace]=[\negthinspace[\mathbb{\mathbb{X}^{\subset}},\mathbb{X}^{\supset}]\negthinspace]$
and $[\negthinspace[\mathbb{Y}]\negthinspace]=[\negthinspace[\mathbb{\mathbb{Y}^{\subset}},\mathbb{Y}^{\supset}]\negthinspace]$.
We define the following operations \cite{DesrochersTAC16} which corresponds
to an interval extension in the sense of Moore \cite{Moore79} applied
to thick sets:
\begin{equation}
\begin{array}{ccc}
[\negthinspace[\mathbb{X}]\negthinspace]\cap[\negthinspace[\mathbb{Y}]\negthinspace] & = & [\negthinspace[\mathbb{\mathbb{X}^{\subset}}\cap\mathbb{Y}^{\subset},\mathbb{X}^{\supset}\cap\mathbb{Y}^{\supset}]\negthinspace]\\{}
[\negthinspace[\mathbb{X}]\negthinspace]\cup[\negthinspace[\mathbb{Y}]\negthinspace] & = & [\negthinspace[\mathbb{\mathbb{X}^{\subset}}\cup\mathbb{Y}^{\subset},\mathbb{X}^{\supset}\cup\mathbb{Y}^{\supset}]\negthinspace]\\
\overline{[\negthinspace[\mathbb{X}]\negthinspace]} & = & [\negthinspace[\overline{\mathbb{X}^{\supset}},\overline{\mathbb{X}^{\subset}}]\negthinspace]\\
\partial[\negthinspace[\mathbb{X}]\negthinspace] & = & [\negthinspace[\mathbb{X}]\negthinspace]\cap\overline{[\negthinspace[\mathbb{X}]\negthinspace]}\\
\mathbf{f}([\negthinspace[\mathbb{X}]\negthinspace]) & = & [\negthinspace[\mathbf{f}(\mathbb{\mathbb{X}^{\subset}}),\mathbf{f}(\mathbb{X}^{\supset})]\negthinspace]\\
\mathbf{f}^{-1}([\negthinspace[\mathbb{X}]\negthinspace]) & = & [\negthinspace[\mathbf{f}^{-1}(\mathbb{\mathbb{X}^{\subset}}),\mathbf{f}^{-1}(\mathbb{X}^{\supset})]\negthinspace]
\end{array}
\end{equation}

We could also have defined these operations as follows: 
\[
\,\begin{array}{ccc}
[\negthinspace[\mathbb{X}]\negthinspace]\diamond[\negthinspace[\mathbb{Y}]\negthinspace] & = & [\negthinspace[\bigcap\left\{ \mathbb{X}\diamond\mathbb{Y},\mathbb{X}\in[\negthinspace[\mathbb{X}]\negthinspace],\mathbb{Y}\in[\negthinspace[\mathbb{Y}]\negthinspace]\right\} ,\bigcup\left\{ \mathbb{X}\diamond\mathbb{Y},\mathbb{X}\in[\negthinspace[\mathbb{X}]\negthinspace],\mathbb{Y}\in[\negthinspace[\mathbb{Y}]\negthinspace]\right\} ]\negthinspace]\end{array}
\]
where $\diamond$ is a binary operator on sets such as $\cup,\cap,\backslash,\dots$,
and 
\[
\phi[\negthinspace[\mathbb{X}]\negthinspace]=[\negthinspace[\bigcap\left\{ \phi(\mathbb{X}),\mathbb{X}\in[\negthinspace[\mathbb{X}]\negthinspace]\right\} ,\bigcup\left\{ \phi(\mathbb{X}),\mathbb{X}\in[\negthinspace[\mathbb{X}]\negthinspace]\right\} ]\negthinspace]
\]
where $\phi$ is a unary operator on sets such as $\partial\mathbb{X},\overline{\mathbb{X}},\dots$

Some of these operations are illustrated by Figure \ref{fig:intersect:thicksets}.
The first subfigure represents the thick set $[\negthinspace[\mathbb{X}]\negthinspace]=[\negthinspace[\mathbb{\mathbb{X}^{\subset}},\mathbb{X}^{\supset}]\negthinspace]$.
The lower bound $\mathbb{\mathbb{X}^{\subset}}$ is painted red; the
penumbra $\mathbb{X}^{\supset}\backslash\mathbb{X}^{\subset}$ is
painted orange; and all $\mathbf{x}$ outside the upper bound $\mathbb{X}^{\supset}$
of $[\negthinspace[\mathbb{X}]\negthinspace]$ is in blue.

\begin{figure}
\begin{centering}
\includegraphics[width=0.5\columnwidth]{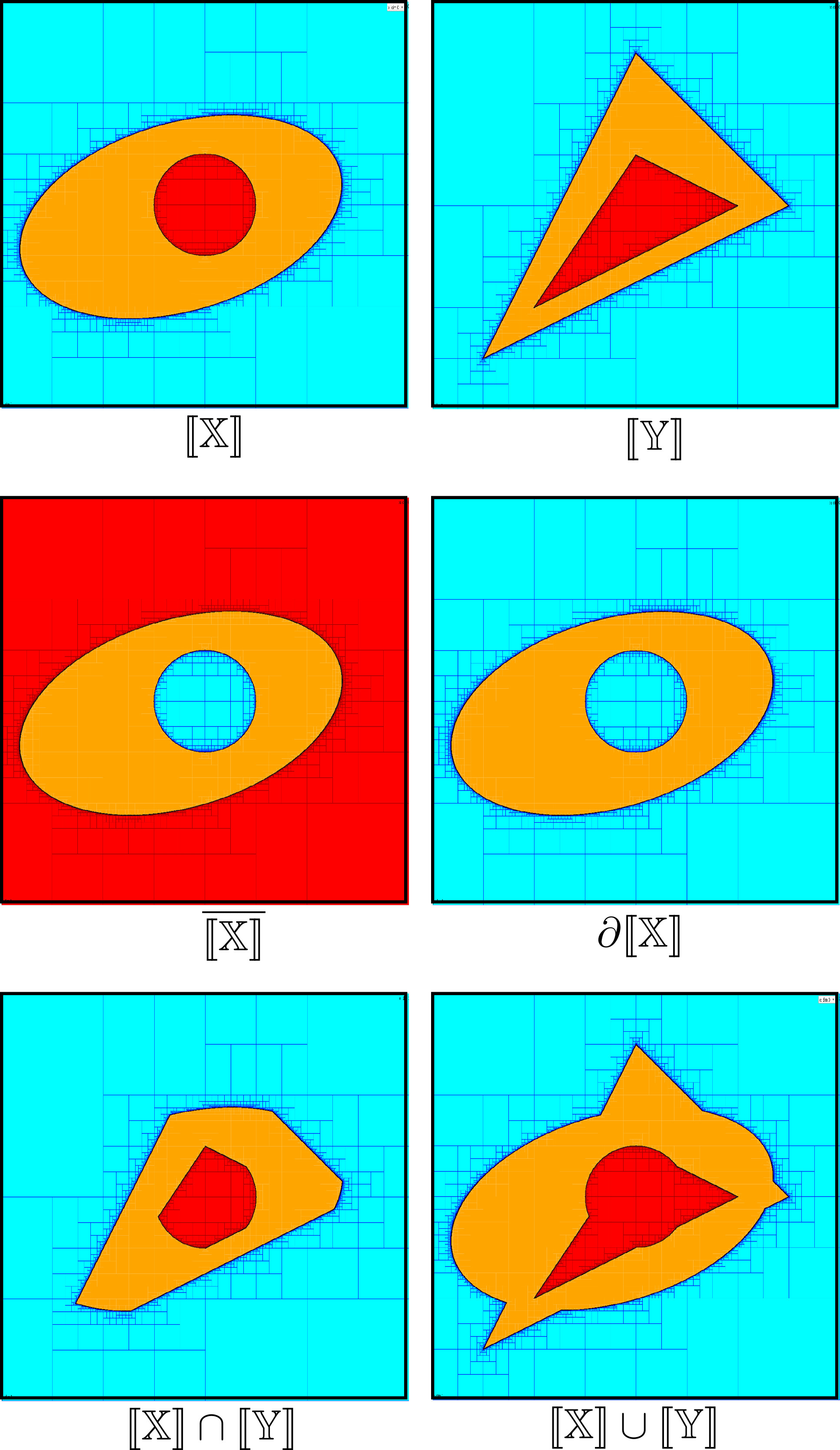}
\par\end{centering}
\caption{Operations between two thick sets. Red means \emph{inside}, Blue means
\emph{outside} and Orange is for the \emph{penumbra}}
\label{fig:intersect:thicksets}
\end{figure}

\subsection{Atoms\label{subsec:Atoms}}

To compute with thick sets, we first need to generate elementary thick
sets, called the \emph{atoms}. They can be given by geometrical sets
such as boxes, polygons or disks. They can also constructed from an
expression of the form $f(\mathbf{x},\mathbf{p})\leq0$, where $f:\mathbb{R}^{n}\times\mathbb{R}^{m}\rightarrow\mathbb{R}$
and $\mathbf{p}\in[\mathbf{p}]$ is the \emph{parameter vector}. This
can be done by the following operator 
\[
[\negthinspace[\sigma]\negthinspace](f,[\mathbf{p}])=[\negthinspace[\mathbb{\mathbb{X}^{\subset}},\mathbb{X}^{\supset}]\negthinspace]
\]

with
\[
\begin{array}{ccc}
\mathbb{\mathbb{X}^{\subset}} & = & \{\mathbf{x}|\forall\mathbf{p}\in[\mathbf{p}]f(\mathbf{x},\mathbf{p})\leq0\}\\
\mathbb{X}^{\supset} & = & \{\mathbf{x}|\exists\mathbf{p}\in[\mathbf{p}]f(\mathbf{x},\mathbf{p})\leq0\}.
\end{array}
\]

\section{Characterizing the sliding surface of an uncertain controlled child
swing \label{sec:application}}

In this section, we propose an original academical example which illustrates
how the thick set algebra can be used to characterize sliding surfaces
of uncertain hybrid systems.

\subsection{Pendulum}

Consider a pendulum depicted in Figure \ref{fig:pendulesimple_x1x2},
left. Its state representation is 
\[
\begin{array}{lll}
\left(\begin{array}{c}
\dot{x}_{1}\\
\dot{x}_{2}
\end{array}\right) & = & \left(\begin{array}{c}
x_{2}\\
-\sin x_{1}+p_{1}+u
\end{array}\right)\end{array}
\]

\begin{figure}[t]
\centering\includegraphics[width=0.8\textwidth]{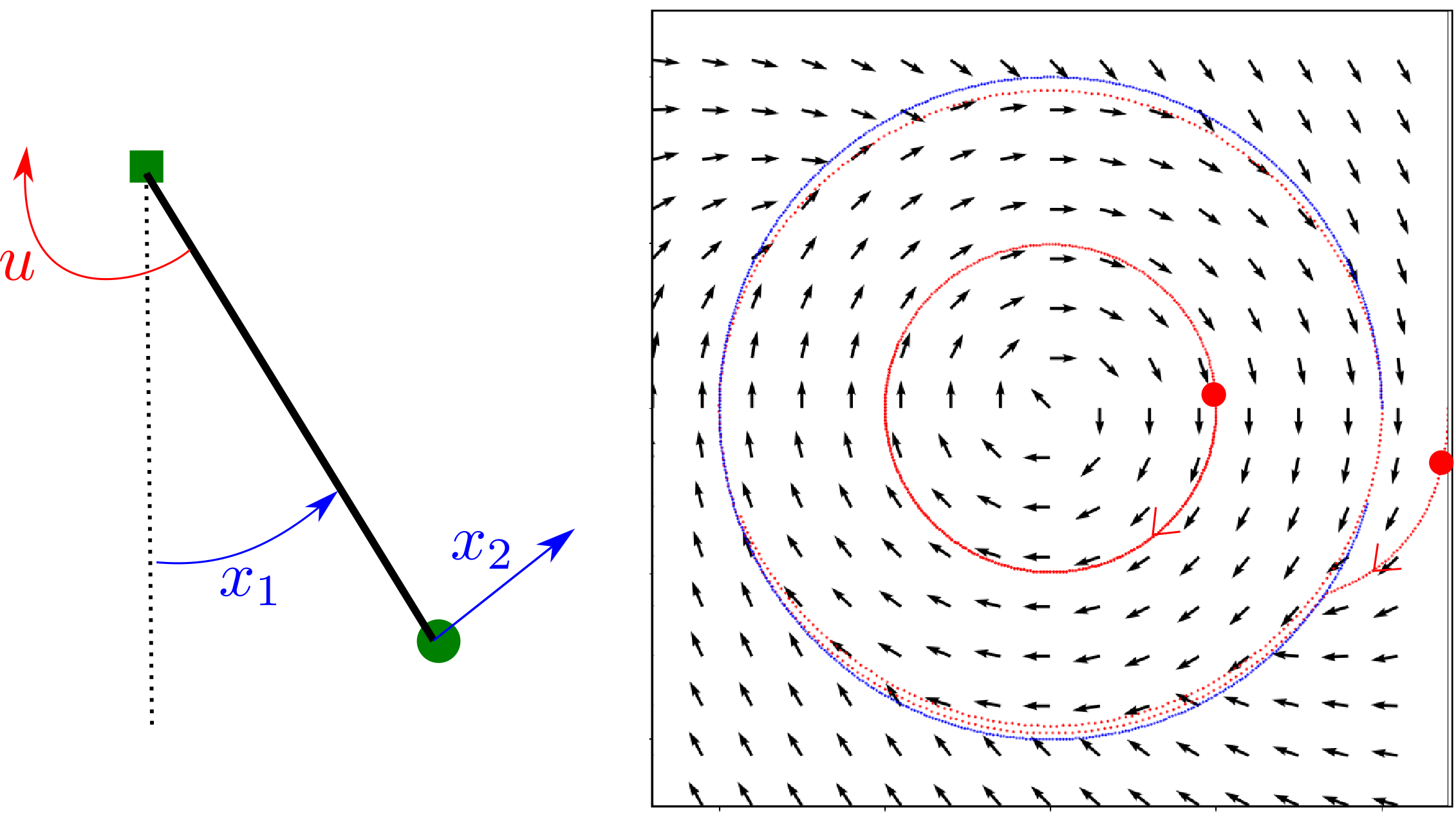}\caption{Child swing with state vector $\mathbf{x}=\left(x_{1},x_{2}\right)$}

\label{fig:pendulesimple_x1x2}
\end{figure}

The pendulum represents a child swing, $u$ is a security break and
$p_{1}$ is a perturbation which can be associated to the force generated
by the child. In a nominal behavior, we have, $u=0$ but when the
energy of the swing becomes too large, the controller generates a
friction $u=-x_{2}$ to slow down the swing. More precisely, the following
controller is:
\[
\text{ if }c(\mathbf{x})<0\text{ then }u=0\text{ else \ensuremath{u=-x_{2}},}
\]
where $c(\mathbf{x})=x_{1}^{2}+x_{2}^{2}-1$ plays the role of the
energy. The corresponding vector field is represented in Figure \ref{fig:pendulesimple_x1x2},
right with two different trajectories (red) for $p_{1}=0$. 

\subsection{Thick approximation of the sliding surface}

Assume that, when the controller alternates indefinitely between $u=0$
(break off) and $u=-x_{2}$ (break on), the controller may be damaged.
We want to compute the sliding states which can be difficult to detect
with simulations.

We have the two fields
\[
\mathbf{f}_{a}=\left(\begin{array}{c}
x_{2}\\
p_{1}-\sin x_{1}
\end{array}\right),\mathbf{f}_{b}=\left(\begin{array}{c}
x_{2}\\
p_{1}-\sin x_{1}-x_{2}
\end{array}\right)
\]
where $p_{1}\in[p_{1}]$ is an uncertain variable. We have

\begin{equation}
\begin{array}{ccl}
\mathcal{L}_{a}^{c}\left(\mathbf{x}\right) & = & \frac{dc}{d\mathbf{x}}\left(\mathbf{x}\right)\cdot\mathbf{f}_{a}\left(\mathbf{x}\right)\\
 & = & \left(\begin{array}{cc}
2x_{1} & 2x_{2}\end{array}\right)\cdot\left(\begin{array}{c}
x_{2}\\
p_{1}-\sin x_{1}
\end{array}\right)\\
 & = & 2x_{1}x_{2}+2x_{2}(p_{1}-\sin x_{1})
\end{array}
\end{equation}
and
\[
\begin{array}{cll}
\mathcal{L}_{b}^{c}\left(\mathbf{x}\right) & = & \frac{dc}{d\mathbf{x}}\left(\mathbf{x}\right)\cdot\mathbf{f}_{b}\left(\mathbf{x}\right)\\
 & = & \left(\begin{array}{cc}
2x_{1} & 2x_{2}\end{array}\right)\cdot\left(\begin{array}{c}
x_{2}\\
p_{1}-\sin x_{1}-x_{2}
\end{array}\right)\\
 & = & 2x_{1}x_{2}+2x_{2}(p_{1}-\sin x_{1}-x_{2})
\end{array}
\]

To take into account the fact that $x_{1}$ and $x_{2}$ are given
to the controller with a given bounded error $p_{2}\in[p_{2}],p_{3}\in[p_{3}]$,
we take 

\[
\mathbb{A}=\left\{ \mathbf{x}|\left(x_{1}+p_{2}\right)^{2}+\left(x_{2}+p_{3}\right)^{2}-1\leq0\right\} .
\]

To compute the paving associated to the expression (\ref{eq:SA}),
we need to build the atoms $[\negthinspace[\mathbb{A}]\negthinspace],[\negthinspace[\mathbb{L}_{a}^{c}]\negthinspace],[\negthinspace[\mathbb{L}_{b}^{c}]\negthinspace]$
as explained in Subsection \ref{subsec:Atoms}. 

If we choose $[p_{i}]=[-0.1,0.1],\forall i$, we get the approximation
represented by Figure \ref{fig:thickset}, Left. It is obtained by
the following statements
\[
\begin{array}{ccl}
1 & \,\, & [p_{1}]:=[p_{2}]:=[p_{3}]:=[-0.1,0.1]\\
2 &  & [\negthinspace[\mathbb{L}_{a}^{c}]\negthinspace]:=[\negthinspace[\sigma]\negthinspace](2x_{1}x_{2}+2x_{2}(p_{1}-\sin x_{1}),[p_{1}])\\
3 &  & [\negthinspace[\mathbb{L}_{b}^{c}]\negthinspace]:=[\negthinspace[\sigma]\negthinspace](2x_{1}x_{2}+2x_{2}(p_{1}-\sin x_{1}-x_{2}),[p_{1}])\\
4 &  & [\negthinspace[\mathbb{A}]\negthinspace]:=[\negthinspace[\sigma]\negthinspace](\left(x_{1}+p_{2}\right)^{2}+\left(x_{2}+p_{3}\right)^{2}-1,[p_{2}]\times[p_{3}])\\
5 &  & [\negthinspace[\mathbb{S}]\negthinspace]:=\partial[\negthinspace[\mathbb{A}]\negthinspace]\cap\overline{[\negthinspace[\mathbb{L}_{a}^{c}]\negthinspace]}\,\cap[\negthinspace[\mathbb{L}_{b}^{c}]\negthinspace]
\end{array}
\]
We note that $\mathbb{S}^{\subset}=\emptyset$, which is consistent
with the fact that the surface to be approximated has no volume. This
is the reason why the outer approximation $\mathbb{S}^{\supset}$
of the surface $\mathbb{S}$ corresponds to the penumbra of the thick
set $[\negthinspace[\mathbb{S}]\negthinspace]$. 

\begin{figure}
\centering\includegraphics[width=0.8\textwidth]{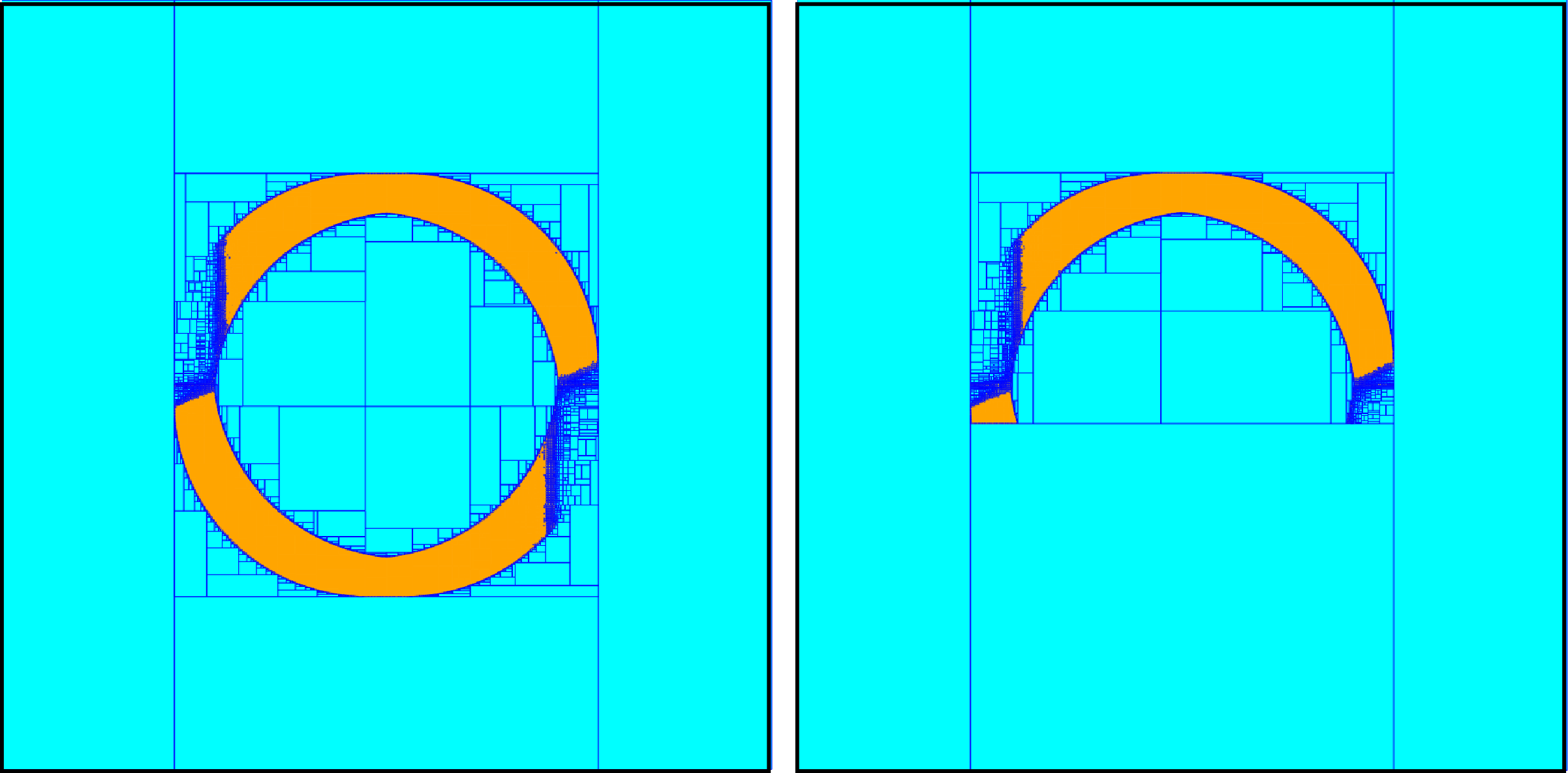}

\caption{The orange zone corresponds to the outer approximation $\mathbb{S}^{\supset}$
of the sliding surface $\mathbb{S}$}
\label{fig:thickset}
\end{figure}

\subsubsection{One more condition}

We extend our example by adding another condition in order to illustrate
the facility of using thick set algebra to compute with uncertain
sets. 

To avoid to frighten the child on the swing, we want that when the
swing goes strongly backward, the break does not switch on. More precisely,
the controller becomes
\[
\text{ if }\underset{c_{1}(\mathbf{x})}{\underbrace{x_{1}^{2}+x_{2}^{2}-1\leq0}}\leq0\text{ or \ensuremath{\underset{c_{2}(\mathbf{x})}{\underbrace{\ensuremath{x_{2}+0.2}}}}\ensuremath{\ensuremath{\leq}0} then }u=0\text{ else \ensuremath{u=-x_{2}}.}
\]
 The \emph{or} condition can be treated using Proposition \ref{prop:algebra}.\textcolor{magenta}{{}
}We have

\begin{equation}
\begin{array}{ccc}
\mathcal{L}_{a}^{c_{2}}\left(\mathbf{x}\right) & = & \frac{dc_{2}}{d\mathbf{x}}\left(\mathbf{x}\right)\cdot\mathbf{f}_{a}\left(\mathbf{x}\right)=p_{1}-\sin x_{1}\end{array}
\end{equation}
and
\begin{equation}
\begin{array}{ccc}
\mathcal{L}_{b}^{c_{2}}\left(\mathbf{x}\right) & = & \frac{dc_{2}}{d\mathbf{x}}\left(\mathbf{x}\right)\cdot\mathbf{f}_{b}\left(\mathbf{x}\right)=p_{1}-\sin x_{1}-x_{2}\end{array}
\end{equation}

An enclosure $[\negthinspace[\mathbb{S}]\negthinspace]$ for the sliding
surface $\mathbb{S}$ is represented by Figure \ref{fig:thickset},
Right. It is obtained by the following statements
\[
\begin{array}{ccl}
1 & \,\, & [p_{1}]:=[p_{2}]:=[p_{3}]:=[-0.1,0.1]\\
2 &  & [\negthinspace[\mathbb{L}_{a}^{c_{1}}]\negthinspace]:=[\negthinspace[\sigma]\negthinspace](2x_{1}x_{2}+2x_{2}(p_{1}-\sin x_{1}),[p_{1}])\\
3 &  & [\negthinspace[\mathbb{L}_{b}^{c_{1}}]\negthinspace]:=[\negthinspace[\sigma]\negthinspace](2x_{1}x_{2}+2x_{2}(p_{1}-\sin x_{1}-x_{2}),[p_{1}])\\
4 &  & [\negthinspace[\mathbb{L}_{a}^{c_{2}}]\negthinspace]:=[\negthinspace[\sigma]\negthinspace](p_{1}-\sin x_{1},[p_{1}])\\
5 &  & [\negthinspace[\mathbb{L}_{b}^{c_{2}}]\negthinspace]:=[\negthinspace[\sigma]\negthinspace](p_{1}-\sin x_{1}-x_{2},[p_{1}])\\
6 &  & [\negthinspace[\mathbb{A}_{1}]\negthinspace]:=[\negthinspace[\sigma]\negthinspace](\left(x_{1}+p_{2}\right)^{2}+\left(x_{2}+p_{3}\right)^{2}-1,[p_{2}]\times[p_{3}])\\
7 &  & [\negthinspace[\mathbb{A}_{2}]\negthinspace]:=[\negthinspace[\sigma]\negthinspace](x_{2}+0.2+p_{3},[p_{3}])\\
8 &  & [\negthinspace[\mathbb{S}_{1}]\negthinspace]:=\partial[\negthinspace[\mathbb{A}_{1}]\negthinspace]\cap\overline{[\negthinspace[\mathbb{L}_{a}^{c_{1}}]\negthinspace]}\,\cap[\negthinspace[\mathbb{L}_{b}^{c_{1}}]\negthinspace]\\
9 &  & [\negthinspace[\mathbb{S}_{2}]\negthinspace]:=\partial[\negthinspace[\mathbb{A}_{2}]\negthinspace]\cap\overline{[\negthinspace[\mathbb{L}_{a}^{c_{2}}]\negthinspace]}\,\cap[\negthinspace[\mathbb{L}_{b}^{c_{2}}]\negthinspace]\\
10 &  & [\negthinspace[\mathbb{S}]\negthinspace]:=\left([\negthinspace[\mathbb{S}_{1}]\negthinspace]\cap\overline{[\negthinspace[\mathbb{A}_{2}]\negthinspace]}\right)\cup\left([\negthinspace[\mathbb{S}_{2}]\negthinspace]\cap\overline{[\negthinspace[\mathbb{A}_{1}]\negthinspace]}\right)
\end{array}
\]

A \textsc{Python} program associated with this example can be tested
here:
\begin{center}
\href{https://www.ensta-bretagne.fr/jaulin/swing.html}{www.ensta-bretagne.fr/jaulin/swing.html}
\par\end{center}

\section{Conclusion\label{sec:Conclusion}}

In this paper, we have presented a new approach based on thick sets
to enclose the sliding surfaces of a cyber-physical system in case
of interval uncertainties. If the state of the system is on this sliding
approximation, it may hesitate indefinitely between two different
strategies. As a result, the system may be trapped on the sliding
surface and the system may be damaged. It is thus important to compute
an approximation of the sliding surface.

An expression given with thick sets can also be written as a quantified
constraint satisfaction problem \cite{Ratschan:01a}. The main advantage
of using a thick set expressions is that it corresponds to an interval
extension of the thin set expression we want to compute. Thick sets
is thus an interval representation of the uncertainties attached to
the set we want to compute. Thick set arithmetic can also be interpreted
as an interval arithmetic used to compute with uncertain subsets of
$\mathbb{R}^{n}$, where as the classical interval arithmetic is used
to compute with uncertain real numbers.

\nocite{*}
\bibliographystyle{eptcs}
\bibliography{swing}

\begin{thebibliography}{10}
\providecommand{\bibitemdeclare}[2]{}
\providecommand{\surnamestart}{}
\providecommand{\surnameend}{}
\providecommand{\urlprefix}{Available at }
\providecommand{\url}[1]{\texttt{#1}}
\providecommand{\href}[2]{\texttt{#2}}
\providecommand{\urlalt}[2]{\href{#1}{#2}}
\providecommand{\doi}[1]{doi:\urlalt{http://dx.doi.org/#1}{#1}}
\providecommand{\bibinfo}[2]{#2}

\bibitemdeclare{article}{Asarin07}
\bibitem{Asarin07}
\bibinfo{author}{E.~\surnamestart Asarin\surnameend},
  \bibinfo{author}{T.~\surnamestart Dang\surnameend} \&
  \bibinfo{author}{A.~\surnamestart Girard\surnameend} (\bibinfo{year}{2007}):
  \emph{\bibinfo{title}{Hybridization methods for the analysis of non-linear
  systems}}.
\newblock {\sl \bibinfo{journal}{Acta Informatica}}
  \bibinfo{volume}{7}(\bibinfo{number}{43}), pp. \bibinfo{pages}{451--476},
  \doi{10.1007/s00236-006-0035-7}.

\bibitemdeclare{inproceedings}{Asarin02}
\bibitem{Asarin02}
\bibinfo{author}{E.~\surnamestart Asarin\surnameend},
  \bibinfo{author}{T.~\surnamestart Dang\surnameend} \&
  \bibinfo{author}{O.~\surnamestart Maler\surnameend} (\bibinfo{year}{2002}):
  \emph{\bibinfo{title}{The d/dt tool for verification of hybrid systems}}.
\newblock In: {\sl \bibinfo{booktitle}{In International Conference on Computer
  Aided Verification}}, \bibinfo{publisher}{Springer}, pp.
  \bibinfo{pages}{365--370}.

\bibitemdeclare{book}{Blanchini08}
\bibitem{Blanchini08}
\bibinfo{author}{F.~\surnamestart Blanchini\surnameend} \&
  \bibinfo{author}{S.~\surnamestart Miani\surnameend} (\bibinfo{year}{2007}):
  \emph{\bibinfo{title}{Set-{Theoretic} {Methods} in {Control}}}.
\newblock \bibinfo{publisher}{Springer Science \& Business Media},
  \doi{10.1007/978-3-319-17933-9}.

\bibitemdeclare{inproceedings}{Bouissou14}
\bibitem{Bouissou14}
\bibinfo{author}{O.~\surnamestart Bouissou\surnameend},
  \bibinfo{author}{A.~\surnamestart Chapoutot\surnameend},
  \bibinfo{author}{A.~\surnamestart Djaballah\surnameend} \&
  \bibinfo{author}{M.~\surnamestart Kieffer\surnameend} (\bibinfo{year}{2014}):
  \emph{\bibinfo{title}{Computation of parametric barrier functions for
  dynamical systems using interval analysis}}.
\newblock In: {\sl \bibinfo{booktitle}{2014 {IEEE} 53rd {Annual} {Conference}
  on {Decision} and {Control} ({CDC})}}, pp. \bibinfo{pages}{753--758},
  \doi{10.1109/CDC.2014.7039472}.

\bibitemdeclare{article}{redaTSF20}
\bibitem{redaTSF20}
\bibinfo{author}{R.~\surnamestart Boukezzoula\surnameend},
  \bibinfo{author}{L.~\surnamestart Jaulin\surnameend},
  \bibinfo{author}{B.~\surnamestart Desrochers\surnameend} \&
  \bibinfo{author}{D.~\surnamestart Coquin\surnameend} (\bibinfo{year}{2020}):
  \emph{\bibinfo{title}{Thick Fuzzy Sets and Their Potential Use in Uncertain
  Fuzzy Computations and Modeling}}.
\newblock {\sl \bibinfo{journal}{IEEE Transactions on Fuzzy Systems}},
  \doi{10.1109/TFUZZ.2020.3018550}.

\bibitemdeclare{inproceedings}{Brefort14}
\bibitem{Brefort14}
\bibinfo{author}{Q.~\surnamestart Brefort\surnameend},
  \bibinfo{author}{L.~\surnamestart Jaulin\surnameend},
  \bibinfo{author}{M.~\surnamestart Ceberio\surnameend} \&
  \bibinfo{author}{V.~\surnamestart Kreinovich\surnameend}
  (\bibinfo{year}{2014}): \emph{\bibinfo{title}{If we take into account that
  constraints are soft,then processing constraints Becomes algorithmically
  solvable}}.
\newblock In: {\sl \bibinfo{booktitle}{Proceedings of the IEEE Series of
  Symposia on Computational Intelligence SSCI'2014}},
  \bibinfo{publisher}{Orlando, Florida, December 9-12},
  \doi{10.1109/CIES.2014.7011823}.

\bibitemdeclare{inproceedings}{Cousot97}
\bibitem{Cousot97}
\bibinfo{author}{P.~\surnamestart Cousot\surnameend} \&
  \bibinfo{author}{R.~\surnamestart Cousot\surnameend} (\bibinfo{year}{1977}):
  \emph{\bibinfo{title}{Abstract Interpretation: A Unified Lattice Model for
  Static Analysis of Programs by Construction or Approximation of Fixpoints}}.
\newblock In: {\sl \bibinfo{booktitle}{Conference Record of the Fourth ACM
  Symposium on Principles of Programming Languages}}, \bibinfo{address}{Los
  Angeles, California}, pp. \bibinfo{pages}{238--252}.

\bibitemdeclare{inproceedings}{Delanoue:attraction:06}
\bibitem{Delanoue:attraction:06}
\bibinfo{author}{N.~\surnamestart Delanoue\surnameend},
  \bibinfo{author}{L.~\surnamestart Jaulin\surnameend} \&
  \bibinfo{author}{B.~\surnamestart Cottenceau\surnameend}
  (\bibinfo{year}{2006}): \emph{\bibinfo{title}{Attraction domain of a
  nonlinear system using interval analysis}}.
\newblock In: {\sl \bibinfo{booktitle}{Twelfth International Conference on
  Principles and Practice of Constraint Programming (IntCP 2006)}},
  \bibinfo{address}{France, Nantes}, pp. \bibinfo{pages}{181--189}.

\bibitemdeclare{article}{DesrochersTAC16}
\bibitem{DesrochersTAC16}
\bibinfo{author}{B.~\surnamestart Desrochers\surnameend} \&
  \bibinfo{author}{L.~\surnamestart Jaulin\surnameend} (\bibinfo{year}{2017}):
  \emph{\bibinfo{title}{Computing a guaranteed approximation the zone explored
  by a robot}}.
\newblock {\sl \bibinfo{journal}{IEEE Transaction on Automatic Control}}
  \bibinfo{volume}{62}(\bibinfo{number}{1}), pp. \bibinfo{pages}{425--430},
  \doi{10.1109/TAC.2016.2530719}.

\bibitemdeclare{article}{DesrochersThick2016}
\bibitem{DesrochersThick2016}
\bibinfo{author}{B.~\surnamestart Desrochers\surnameend} \&
  \bibinfo{author}{L.~\surnamestart Jaulin\surnameend} (\bibinfo{year}{2017}):
  \emph{\bibinfo{title}{{Thick set inversion}}}.
\newblock {\sl \bibinfo{journal}{Artifical Intelligence}}
  \bibinfo{volume}{249}, pp. \bibinfo{pages}{1--18},
  \doi{10.1016/j.artint.2017.04.004}.

\bibitemdeclare{inproceedings}{desrochers:iros2015}
\bibitem{desrochers:iros2015}
\bibinfo{author}{B.~\surnamestart Desrochers\surnameend},
  \bibinfo{author}{S.~\surnamestart Lacroix\surnameend} \&
  \bibinfo{author}{L.~\surnamestart Jaulin\surnameend} (\bibinfo{year}{2015}):
  \emph{\bibinfo{title}{Set-Membership Approach to the Kidnapped Robot
  Problem}}.
\newblock In: {\sl \bibinfo{booktitle}{IROS 2015}},
  \doi{10.1109/IROS.2015.7353897}.

\bibitemdeclare{article}{Drakunov92}
\bibitem{Drakunov92}
\bibinfo{author}{S.~\surnamestart Drakunov\surnameend} \&
  \bibinfo{author}{V.~\surnamestart Utkin\surnameend} (\bibinfo{year}{1992}):
  \emph{\bibinfo{title}{Sliding mode control in dynamic systems}}.
\newblock {\sl \bibinfo{journal}{International Journal of Control}}
  \bibinfo{volume}{55}(\bibinfo{number}{4}), pp. \bibinfo{pages}{1029--1037},
  \doi{10.1016/0005-1098(76)90076-5}.

\bibitemdeclare{inbook}{dubois:jaulin:prade}
\bibitem{dubois:jaulin:prade}
\bibinfo{author}{D.~\surnamestart Dubois\surnameend},
  \bibinfo{author}{L.~\surnamestart Jaulin\surnameend} \&
  \bibinfo{author}{H.~\surnamestart Prade\surnameend} (\bibinfo{year}{2020}):
  \emph{\bibinfo{title}{Thick Sets, Multiple-Valued Mappings and Possibility
  Theory}}, pp. \bibinfo{pages}{101--109}.
\newblock \bibinfo{publisher}{Springer}.

\bibitemdeclare{article}{Frehse:08}
\bibitem{Frehse:08}
\bibinfo{author}{G.~\surnamestart Frehse\surnameend} (\bibinfo{year}{2008}):
  \emph{\bibinfo{title}{PHAVer: Algorithmic Verification of Hybrid Systems}}.
\newblock {\sl \bibinfo{journal}{International Journal on Software Tools for
  Technology Transfer}} \bibinfo{volume}{10}(\bibinfo{number}{3}), pp.
  \bibinfo{pages}{23--48}, \doi{10.1007/s10009-007-0062-x}.

\bibitemdeclare{inproceedings}{Goubault06staticanalysis}
\bibitem{Goubault06staticanalysis}
\bibinfo{author}{E.~\surnamestart Goubault\surnameend} \&
  \bibinfo{author}{S.~\surnamestart Putot\surnameend} (\bibinfo{year}{2006}):
  \emph{\bibinfo{title}{Static Analysis of Numerical Algorithms}}.
\newblock In: {\sl \bibinfo{booktitle}{In Proceedings of SAS 06, LNCS 4134}},
  \bibinfo{publisher}{Springer-Verlag}, pp. \bibinfo{pages}{18--34}.

\bibitemdeclare{article}{jaulin:sliding:20}
\bibitem{jaulin:sliding:20}
\bibinfo{author}{L.~\surnamestart Jaulin\surnameend} \& \bibinfo{author}{F.~Le
  \surnamestart Bars\surnameend} (\bibinfo{year}{2020}):
  \emph{\bibinfo{title}{Characterizing sliding surfaces of cyber-physical
  systems}}.
\newblock {\sl \bibinfo{journal}{Acta Cybernetica}} \bibinfo{volume}{24}, pp.
  \bibinfo{pages}{431--448}, \doi{10.4467/20838476SI.11.001.0287}.

\bibitemdeclare{inproceedings}{taha:13:zeno}
\bibitem{taha:13:zeno}
\bibinfo{author}{M.~\surnamestart Konecny\surnameend},
  \bibinfo{author}{W.~\surnamestart Taha\surnameend},
  \bibinfo{author}{J.~\surnamestart Duracz\surnameend},
  \bibinfo{author}{A.~\surnamestart Duracz\surnameend} \&
  \bibinfo{author}{A.~\surnamestart Ames\surnameend} (\bibinfo{year}{2013}):
  \emph{\bibinfo{title}{Enclosing the behavior of a hybrid system up to and
  beyond a Zeno point}}.
\newblock In: {\sl \bibinfo{booktitle}{Cyber-Physical Systems, Networks, and
  Applications (CPSNA)}}, \doi{10.1109/CPSNA.2013.6614258}.

\bibitemdeclare{article}{Kreinovich:97}
\bibitem{Kreinovich:97}
\bibinfo{author}{V.~\surnamestart Kreinovich\surnameend}, \bibinfo{author}{A.V.
  \surnamestart Lakeyev\surnameend}, \bibinfo{author}{J.~\surnamestart
  Rohn\surnameend} \& \bibinfo{author}{P.T. \surnamestart Kahl\surnameend}
  (\bibinfo{year}{1997}): \emph{\bibinfo{title}{Computational Complexity and
  Feasibility of Data Processing and Interval Computations}}.
\newblock {\sl \bibinfo{journal}{Reliable Computing}}
  \bibinfo{volume}{4}(\bibinfo{number}{4}), pp. \bibinfo{pages}{405--409},
  \doi{10.1007/978-1-4757-2793-7}.

\bibitemdeclare{article}{lemezo:tac:18}
\bibitem{lemezo:tac:18}
\bibinfo{author}{T.~Le \surnamestart M\'ezo\surnameend},
  \bibinfo{author}{L.~\surnamestart Jaulin\surnameend} \&
  \bibinfo{author}{B.~\surnamestart Zerr\surnameend} (\bibinfo{year}{2017}):
  \emph{\bibinfo{title}{An interval approach to compute invariant sets}}.
\newblock {\sl \bibinfo{journal}{IEEE Transaction on Automatic Control}}
  \bibinfo{volume}{62}, pp. \bibinfo{pages}{4236--4243},
  \doi{10.1109/TAC.2017.2685241}.

\bibitemdeclare{inproceedings}{Mitchell07}
\bibitem{Mitchell07}
\bibinfo{author}{I.~\surnamestart Mitchell\surnameend} (\bibinfo{year}{2007}):
  \emph{\bibinfo{title}{Comparing forward and backward reachability as tools
  for safety analysis}}.
\newblock In \bibinfo{editor}{A.~\surnamestart Bemporad\surnameend},
  \bibinfo{editor}{A.~\surnamestart Bicchi\surnameend} \&
  \bibinfo{editor}{G.~\surnamestart Buttazzo\surnameend}, editors: {\sl
  \bibinfo{booktitle}{Hybrid Systems: Computation and Control}},
  \bibinfo{publisher}{Springer-Verlag}, pp. \bibinfo{pages}{428--443},
  \doi{10.1109/4.16303}.

\bibitemdeclare{incollection}{mitchell:validating:2001}
\bibitem{mitchell:validating:2001}
\bibinfo{author}{I.~\surnamestart Mitchell\surnameend},
  \bibinfo{author}{A.~\surnamestart Bayen\surnameend} \&
  \bibinfo{author}{C.~\surnamestart Tomlin\surnameend} (\bibinfo{year}{2001}):
  \emph{\bibinfo{title}{Validating a {Hamilton}-{Jacobi} {Approximation} to
  {Hybrid} {System} {Reachable} {Sets}}}.
\newblock In \bibinfo{editor}{M.~\surnamestart Benedetto\surnameend} \&
  \bibinfo{editor}{A.~\surnamestart Sangiovanni-Vincentelli\surnameend},
  editors: {\sl \bibinfo{booktitle}{Hybrid {Systems}: {Computation} and
  {Control}}}, {\sl \bibinfo{series}{Lecture {Notes} in {Computer} {Science}}}
  \bibinfo{volume}{2034}, \bibinfo{publisher}{Springer Berlin Heidelberg}, pp.
  \bibinfo{pages}{418--432}, \doi{10.1006/jcph.1999.6345}.

\bibitemdeclare{book}{Moore79}
\bibitem{Moore79}
\bibinfo{author}{R.~E. \surnamestart Moore\surnameend} (\bibinfo{year}{1979}):
  \emph{\bibinfo{title}{Methods and {A}pplications of {I}nterval {A}nalysis}}.
\newblock \bibinfo{publisher}{SIAM}, \bibinfo{address}{Philadelphia, PA},
  \doi{10.1137/1.9781611970906}.

\bibitemdeclare{article}{Ramdani:Nedialkov11}
\bibitem{Ramdani:Nedialkov11}
\bibinfo{author}{N.~\surnamestart Ramdani\surnameend} \&
  \bibinfo{author}{N.~\surnamestart Nedialkov\surnameend}
  (\bibinfo{year}{2011}): \emph{\bibinfo{title}{{Computing Reachable Sets for
  Uncertain Nonlinear Hybrid Systems using Interval Constraint Propagation
  Techniques}}}.
\newblock {\sl \bibinfo{journal}{Nonlinear Analysis: Hybrid Systems}}
  \bibinfo{volume}{5}(\bibinfo{number}{2}), pp. \bibinfo{pages}{149--162},
  \doi{10.1016/j.nahs.2010.05.010}.

\bibitemdeclare{article}{Ratschan:01a}
\bibitem{Ratschan:01a}
\bibinfo{author}{S.~\surnamestart Ratschan\surnameend} (\bibinfo{year}{2002}):
  \emph{\bibinfo{title}{Approximate Quantified Constraint Solving by
  Cylindrical Box Decomposition}}.
\newblock {\sl \bibinfo{journal}{Reliable Computing}}
  \bibinfo{volume}{8}(\bibinfo{number}{1}), pp. \bibinfo{pages}{21--42},
  \doi{10.1023/A:1014785518570}.

\bibitemdeclare{article}{Ratschan:She:10}
\bibitem{Ratschan:She:10}
\bibinfo{author}{S.~\surnamestart Ratschan\surnameend} \&
  \bibinfo{author}{Z.~\surnamestart She\surnameend} (\bibinfo{year}{2010}):
  \emph{\bibinfo{title}{{Providing a Basin of Attraction to a Target Region of
  Polynomial Systems by Computation of Lyapunov-like Functions}}}.
\newblock {\sl \bibinfo{journal}{SIAM J. Control and Optimization}}
  \bibinfo{volume}{48}(\bibinfo{number}{7}), pp. \bibinfo{pages}{4377--4394},
  \doi{10.1137/090749955}.

\bibitemdeclare{inproceedings}{Rauh2009IntervalAT}
\bibitem{Rauh2009IntervalAT}
\bibinfo{author}{A.~\surnamestart Rauh\surnameend} \&
  \bibinfo{author}{E.~\surnamestart Auer\surnameend} (\bibinfo{year}{2009}):
  \emph{\bibinfo{title}{Interval Approaches to Reliable Control of Dynamical
  Systems}}.
\newblock In: {\sl \bibinfo{booktitle}{Computer-assisted proofs - tools,
  methods and applications}}.

\bibitemdeclare{article}{rohouAut18}
\bibitem{rohouAut18}
\bibinfo{author}{S.~\surnamestart Rohou\surnameend},
  \bibinfo{author}{L.~\surnamestart Jaulin\surnameend},
  \bibinfo{author}{M.~\surnamestart Mihaylova\surnameend},
  \bibinfo{author}{F.~Le \surnamestart Bars\surnameend} \&
  \bibinfo{author}{S.~\surnamestart Veres\surnameend} (\bibinfo{year}{2018}):
  \emph{\bibinfo{title}{{Reliable non-linear state estimation involving time
  uncertainties}}}.
\newblock {\sl \bibinfo{journal}{Automatica}}, pp. \bibinfo{pages}{379--388},
  \doi{10.1016/j.automatica.2018.03.074}.

\bibitemdeclare{article}{Romig19}
\bibitem{Romig19}
\bibinfo{author}{S.~\surnamestart Romig\surnameend},
  \bibinfo{author}{L.~\surnamestart Jaulin\surnameend} \&
  \bibinfo{author}{A.~\surnamestart Rauh\surnameend} (\bibinfo{year}{2019}):
  \emph{\bibinfo{title}{Using Interval Analysis to Compute the Invariant Set of
  a Nonlinear Closed-Loop Control System}}.
\newblock {\sl \bibinfo{journal}{Algorithms}}
  \bibinfo{volume}{12}(\bibinfo{number}{262}), \doi{10.3390/a12120262}.

\bibitemdeclare{inproceedings}{SaintPierre02}
\bibitem{SaintPierre02}
\bibinfo{author}{P.~\surnamestart Saint-Pierre\surnameend}
  (\bibinfo{year}{2002}): \emph{\bibinfo{title}{Hybrid kernels and capture
  basins for impulse constrained systems}}.
\newblock In \bibinfo{editor}{C.J. \surnamestart Tomlin\surnameend} \&
  \bibinfo{editor}{M.R. \surnamestart Greenstreet\surnameend}, editors: {\sl
  \bibinfo{booktitle}{in Hybrid Systems: Computation and Control}},
  \bibinfo{volume}{2289}, \bibinfo{publisher}{Springer-Verlag}, pp.
  \bibinfo{pages}{378--392}, \doi{10.1007/3-540-48983-5}.

\bibitemdeclare{article}{alexandre:chap:16}
\bibitem{alexandre:chap:16}
\bibinfo{author}{J.~Alexandre~Dit \surnamestart Sandretto\surnameend} \&
  \bibinfo{author}{A.~\surnamestart Chapoutot\surnameend}
  (\bibinfo{year}{2016}): \emph{\bibinfo{title}{Validated Simulation of
  Differential Algebraic Equations with {R}unge-{K}utta Methods}}.
\newblock {\sl \bibinfo{journal}{Reliable Computing}} \bibinfo{volume}{22}.

\bibitemdeclare{inproceedings}{taha:15:acumen}
\bibitem{taha:15:acumen}
\bibinfo{author}{W.~\surnamestart Taha\surnameend} \&
  \bibinfo{author}{A.~\surnamestart Duracz\surnameend} (\bibinfo{year}{2015}):
  \emph{\bibinfo{title}{Acumen: An Open-source Testbed for Cyber-Physical
  Systems Research}}.
\newblock In: {\sl \bibinfo{booktitle}{CYCLONE'15}},
  \doi{10.1007/978-3-319-47063-4\_11}.

\bibitemdeclare{article}{wilczak2011}
\bibitem{wilczak2011}
\bibinfo{author}{D.~\surnamestart Wilczak\surnameend} \&
  \bibinfo{author}{P.~\surnamestart Zgliczynski\surnameend}
  (\bibinfo{year}{2011}): \emph{\bibinfo{title}{{C}r-{L}ohner algorithm}}.
\newblock {\sl \bibinfo{journal}{Schedae Informaticae}} \bibinfo{volume}{20},
  pp. \bibinfo{pages}{9--46}, \doi{10.4467/20838476SI.11.001.0287}.

\end{thebibliography}
\end{document}